\documentclass[journal,10pt]{IEEEtran}
\usepackage{cite}
\usepackage{amsmath,amssymb,amsfonts}
\usepackage{algorithmic}
\usepackage{lipsum,adjustbox}
\usepackage{verbatim}
\usepackage{graphics}
\usepackage{stackengine}
\usepackage{pgfplots}
\pgfplotsset{width=7cm,compat=1.8}
\usepackage{multirow}
\usepackage{soul}
\usepackage{amssymb}
\usepackage{amsmath}
\usepackage{algorithm}
\usepackage{algorithmic}
\usepackage{graphicx}
\usepackage{subfig}
\usepackage{textcomp} 
\usepackage{booktabs}
\usepackage{stfloats}
\usepackage{scalefnt}
\usepackage{mathrsfs}       
\DeclareMathAlphabet{\pazocal}{OMS}{zplm}{m}{n}
\usepackage[nolist]{acronym}
    
 \usepackage{bm}

\newcommand{\figref}[1]{Figure~\ref{#1}}

\newcommand{\compl}{\mathbb{C}}         

\newcommand{\ma}  [1]{ \bm{#1} } 

\newcommand{\set} [1]{{\mathcal {#1}}} 
\newcommand{\Kon} {\set{K}_{\text{on}}} 
\newcommand{\Kp} {\set{K}_{\text{p}}} 
\newcommand{\Kd} {\set{K}_{\text{d}}} 
\newcommand{\Kn} {\set{K}_{\text{n}}} 

\newcommand{\RS} {\set{RS}} 
\newcommand{\URS} {\set{URS}} 

\begin{acronym}
\acro{DSRC}{dedicated short-range communications}
\acro{C-ITS}{cooperative intelligent transport system}
\acro{RSU}{road side unit}
\acro{Bi}{bi-directional}
\acro{TR}{throughput}
\acro{TDL}{tapped delay line}
\acro{SRNN}{Simple RNN}
\acro{ITS}{Intelligent Transportation Systems}
\acro{IEEE}{Institute of Electrical and Electronics Engineers}
\acro{WAVE}{Wireless Access in Vehicular Environment}
\acro{V2V}{vehicle-to-vehicle}
\acro{V2I}{vehicle-to-infrastructure}
\acro{CCH}{control channel}
\acro{SCH}{service channels}
\acro{STS}{short training symbols}
\acro{LTS}{long training symbols}
\acro{SS}{signal symbol}
\acro{SoA}{state-of-the-art}
\acro{DPA}{data-pilot aided}
\acro{STA}{spectral temporal averaging}
\acro{EL}{ensemble learning}
\acro{FL}{frame length}
\acro{CDP}{constructed data pilots}
\acro{TRFI}{time-domain reliable test frequency domain interpolation}
\acro{MMSE-VP}{minimum mean square error using virtual pilots}
\acro{iCDP}{improved CDP}
\acro{SBS}{symbol-by-symbol}
\acro{FBF}{frame-by-frame}
\acro{E-TRFI}{Enhanced TRFI}
\acro{SR-CNN}{super resolution CNN}
\acro{DN-CNN}{denoising CNN}
\acro{RBF}{radial basis function}
\acro{CNN}{convolutional neural network}
\acro{TS-ChannelNet}{temporal spectral ChannelNet}
\acro{WSSUS}{wide-sense stationary uncorrelated scattering}
\acro{TDR}{transmission data rate}
\acro{LSTM}{long short-term memory}
\acro{ALS}{accurate LS}
\acro{SLS}{simple LS}
\acro{ChannelNet}{channel network}
\acro{ADD-TT}{average decision-directed with time truncation}
\acro{WI}{weighted interpolation}
\acro{DD}{decision-directed}
\acro{SR-ConvLSTM}{super resolution convolutional long short-term memory}
\acro{RS}{reliable subcarriers}
\acro{URS}{unreliable subcarriers}
\acro{AE-DNN}{auto-encoder deep neural network}
\acro{AE}{auto-encoder}
\acro{T-DFT}{truncated discrete Fourier transform}
\acro{TA-TDFT}{temporal averaging T-DFT}
\acro{TA}{time averaging}
\acro{PDP}{power delay profile}
\acro{1G}{first generation}
\acro{2G}{second generation}
\acro{3G}{third generation}
\acro{3GPP}{Third Generation Partnership Project}
\acro{4G}{fourth generation}
\acro{5G}{fifth generation}
\acro{802.11}{IEEE 802.11 specifications}
\acro{A/D}{analog-to-digital}
\acro{ADC}{analog-to-digital}
\acro{AM}{amplitude modulation}
\acro{AP}{access point}
\acro{AR}{augmented reality}
\acro{ASIC}{application-specific integrated circuit}
\acro{ASIP}{Application Specific Integrated Processors}
\acro{AWGN}{additive white Gaussian noise}
\acro{BCJR}{Bahl, Cocke, Jelinek and Raviv}
\acro{BER}{bit error rate}
\acro{BFDM}{bi-orthogonal frequency division multiplexing}
\acro{BPSK}{binary phase shift keying}
\acro{BS}{base stations}
\acro{CA}{carrier aggregation}
\acro{CAF}{cyclic autocorrelation function}
\acro{Car-2-x}{car-to-car and car-to-infrastructure communication}
\acro{CAZAC}{constant amplitude zero autocorrelation waveform}
\acro{CB-FMT}{cyclic block filtered multitone}
\acro{CCDF}{complementary cumulative density function}
\acro{CDF}{cumulative density function}
\acro{CDMA}{code-division multiple access}
\acro{CFO}{carrier frequency offset}
\acro{CIR}{channel impulse response}
\acro{CM}{complex multiplication}
\acro{COFDM}{coded-\acs{OFDM}}
\acro{CoMP}{coordinated multi point}
\acro{COQAM}{cyclic OQAM}
\acro{CP}{cyclic prefix}
\acro{CR}{cognitive radio}
\acro{CRC}{cyclic redundancy check}
\acro{CRLB}{Cram\'{e}r-Rao lower bound}
\acro{CS}{cyclic suffix}
\acro{CSI}{channel state information}
\acro{CSMA}{carrier-sense multiple access}
\acro{CWCU}{component-wise conditionally unbiased}
\acro{D/A}{digital-to-analog}
\acro{D2D}{device-to-device}
\acro{DAC}{digital-to-analog}
\acro{DC}{direct current}
\acro{DFE}{decision feedback equalizer}
\acro{DFT}{discrete Fourier transform}
\acro{DL}{deep learning}
\acro{DMT}{discrete multitone}
\acro{DNN}{deep neural network}
\acro{FNN}{feed-forward neural network}
\acro{DSA}{dynamic spectrum access}
\acro{DSL}{digital subscriber line}
\acro{DSP}{digital signal processor}
\acro{DTFT}{discrete-time Fourier transform}
\acro{DVB}{digital video broadcasting}
\acro{DVB-T}{terrestrial digital video broadcasting}
\acro{DWMT}{discrete wavelet multi tone}
\acro{DZT}{discrete Zak transform}
\acro{E2E}{end-to-end}
\acro{eNodeB}{evolved node b base station}
\acro{E-SNR}{effective signal-to-noise ratio}
\acro{EVD}{eigenvalue decomposition}
\acro{FBMC}{filter bank multicarrier}
\acro{FD}{frequency-domain}
\acro{FDD}{frequency-division duplexing}
\acro{FDE}{frequency domain equalization}
\acro{FDM}{frequency division multiplex}
\acro{FDMA}{frequency-division multiple access}
\acro{FEC}{forward error correction}
\acro{FER}{frame error rate}
\acro{FFT}{fast Fourier transform}
\acro{FIR}{finite impulse response}
\acro{FM}		{frequency modulation}
\acro{FMT}{filtered multi tone}
\acro{FO}{frequency offset}
\acro{F-OFDM}{filtered-\acs{OFDM}}
\acro{FPGA}{field programmable gate array}
\acro{FSC}{frequency selective channel}
\acro{FS-OQAM-GFDM}{frequency-shift OQAM-GFDM}
\acro{FT}{Fourier transform}
\acro{FTD}{fractional time delay}
\acro{FTN}{faster-than-Nyquist signaling}
\acro{GFDM}{generalized frequency division multiplexing}
\acro{GFDMA}{generalized frequency division multiple access}
\acro{GMC-CDM}{generalized	multicarrier code-division multiplexing}
\acro{GNSS}{global navigation satellite system}
\acro{GS}{guard symbols}
\acro{GSM}{Groupe Sp\'{e}cial Mobile}
\acro{GUI}{graphical user interface}
\acro{H2H}{human-to-human}
\acro{H2M}{human-to-machine}
\acro{HTC}{human type communication}
\acro{I}{in-phase}
\acro{i.i.d.}{independent and identically distributed}
\acro{IB}{in-band}
\acro{IBI}{inter-block interference}
\acro{IC}{interference cancellation}
\acro{ICI}{inter-carrier interference}
\acro{ICT}{information and communication technologies}
\acro{ICV}{information coefficient vector}
\acro{IDFT}{inverse discrete Fourier transform}
\acro{IDMA}{interleave division multiple access}
\acro{IEEE}{institute of electrical and electronics engineers}
\acro{IF}{intermediate frequency}
\acro{IFFT}{inverse fast Fourier transform}
\acro{IoT}{Internet of Things}
\acro{IOTA}{isotropic orthogonal transform algorithm}
\acro{IP}{internet protocole}
\acro{IP-core}{intellectual property core}
\acro{ISDB-T}{terrestrial integrated services digital broadcasting}
\acro{ISDN}{integrated services digital network}
\acro{ISI}{inter-symbol interference}
\acro{ITU}{International Telecommunication Union}
\acro{IUI}{inter-user interference}
\acro{LAN}{local area netwrok}
\acro{LLR}{log-likelihood ratio}
\acro{LMMSE}{linear minimum mean square error}
\acro{LNA}{low noise amplifier}
\acro{LO}{local oscillator}
\acro{LOS}{line-of-sight}
\acro{LP}{low-pass}
\acro{LPF}{low-pass filter}
\acro{LS}{least squares}
\acro{LTE}{Long Term Evolution}
\acro{LTE-A}{LTE-Advanced}
\acro{LTIV}{linear time invariant}
\acro{LTV}{linear time-variant}
\acro{LUT}{lookup table}
\acro{M2M}{machine-to-machine}
\acro{MA}{multiple access}
\acro{MAC}{multiple access control}
\acro{MAP}{maximum a posteriori}
\acro{MC}{multicarrier}
\acro{MCA}{multicarrier access}
\acro{MCM}{multicarrier modulation}
\acro{MCS}{modulation coding scheme}
\acro{MF}{matched filter}
\acro{MF-SIC}{matched filter with successive interference cancellation}
\acro{MIMO}{multiple-input, multiple-output}
\acro{MISO}{multiple-input single-output}
\acro{ML}{machien learning}
\acro{MLD}{maximum likelihood detection}
\acro{MLE}{maximum likelihood estimator}
\acro{MMSE}{minimum mean squared error}
\acro{MRC}{maximum ratio combining}
\acro{MS}{mobile stations}
\acro{MSE}{mean squared error}
\acro{MSK}{Minimum-shift keying}
\acro{MSSS}[MSSS]	{mean-square signal separation}
\acro{MTC}{machine type communication}
\acro{MU}{multi user}
\acro{MVUE}{minimum variance unbiased estimator}
\acro{NEF}{noise enhancement factor}
\acro{NLOS}{non-line-of-sight}
\acro{NMSE}{normalized mean-squared error}
\acro{NOMA}{non-orthogonal multiple access}
\acro{NPR}{near-perfect reconstruction}
\acro{NRZ}{non-return-to-zero}
\acro{OFDM}{orthogonal frequency division multiplexing}
\acro{OFDMA}{orthogonal frequency division multiple access}
\acro{OOB}{out-of-band}
\acro{OQAM}{offset quadrature amplitude modulation}
\acro{OQPSK}{offset quadrature phase shift keying}
\acro{OTFS}{orthogonal time frequency space}
\acro{PA}{power amplifier}
\acro{PAM}{pulse amplitude modulation}
\acro{PAPR}{peak-to-average power ratio}
\acro{PC-CC}{parallel concatenated convolutional code}
\acro{PCP}{pseudo-circular pre/post-amble}
\acro{PD}{probability of detection}
\acro{pdf}{probability density function}
\acro{PDF}{probability distribution function}

\acro{PFA}{probability of false alarm}
\acro{PHY}{physical layer}
\acro{PIC}{parallel interference cancellation}
\acro{PLC}{power line communication}
\acro{PMF}{probability mass function}
\acro{PN}{pseudo noise}
\acro{ppm}{parts per million}
\acro{PRB}{physical resource block}
\acro{PRB}{physical resource block}
\acro{PSD}{power spectral density}
\acro{Q}{quadrature-phase}
\acro{QAM}{quadrature amplitude modulation}
\acro{QoS}{quality of service}
\acro{QPSK}{quadrature phase shift keying}
\acro{R/W}{read-or-write}
\acro{RAM}{random-access memmory}
\acro{RAN}{radio access network}
\acro{RAT}{radio access technologies}
\acro{RC}{raised cosine}
\acro{RF}{radio frequency}
\acro{rms}{root mean square}
\acro{RRC}{root raised cosine}
\acro{RW}{read-and-write}
\acro{SC}{single-carrier}
\acro{SCA}{single-carrier access}
\acro{SC-FDE}{single-carrier with frequency domain equalization}
\acro{SC-FDM}{single-carrier frequency division multiplexing}
\acro{SC-FDMA}{single-carrier frequency division multiple access}
\acro{SD}{sphere decoding}
\acro{SDD}{space-division duplexing}
\acro{SDMA}{space division multiple access}
\acro{SDR}{software-defined radio}
\acro{SDW}{software-defined waveform}
\acro{SEFDM}{spectrally efficient frequency division multiplexing}
\acro{SE-FDM}{spectrally efficient frequency division multiplexing}
\acro{SER}{symbol error rate}
\acro{SIC}{successive interference cancellation}
\acro{SINR}{signal-to-interference-plus-noise ratio}
\acro{SIR}{signal-to-interference ratio}
\acro{SISO}{single-input, single-output}
\acro{SMS}{Short Message Service}
\acro{SNR}{signal-to-noise ratio}
\acro{STC}{space-time coding}
\acro{STFT}{short-time Fourier transform}
\acro{STO}{symbol time offset}
\acro{SU}{single user}
\acro{AI}{Artificial intelligence}
\acro{SVD}{singular value decomposition}
\acro{TD}{time-domain}	
\acro{TDD}{time-division duplexing}
\acro{TDMA}{time-division multiple access}
\acro{TFL}{time-frequency localization}
\acro{TO}{time offset}
\acro{TS-OQAM-GFDM}{time-shifted OQAM-GFDM}
\acro{XAI}{explainable AI}
\acro{UE}{user equipment}
\acro{VTV-US}{Vehicle-to-Vehicle urban scenario}
\acro{VTI-US}{Vehicle-to-Infrastructure urban scenario}
\acro{UFMC}{universally filtered multicarrier}
\acro{UL}{uplink}
\acro{US}{uncorrelated scattering}
\acro{USB}{universal serial bus}
\acro{UW}{unique word}
\acro{VLC}{visible light communications}
\acro{VR}{virtual reality}
\acro{WCP}{windowing and \acs{CP}}	\acro{ANN}{artificial neural network}
\acro{GRU}{gated recurrent unit}
\acro{RNN}{recurrent neural network}
\acro{WHT}{Walsh-Hadamard transform}
\acro{WiMAX}{worldwide interoperability for microwave access}
\acro{WLAN}{wireless local area network}
\acro{W-OFDM}{windowed-\acs{OFDM}}	
\acro{WOLA}{windowing and overlapping}	
\acro{WSS}{wide-sense stationary}
\acro{ZCT}{Zadoff-Chu transform}
\acro{ZF}{zero-forcing}
\acro{ZMCSCG}{zero-mean circularly-symmetric complex Gaussian}
\acro{ZP}{zero-padding}
\acro{ZT}{zero-tail}
\end{acronym}

\usepackage[utf8]{inputenc}
\usepackage[english]{babel}
\usepackage{amsthm}
\usepackage{commath}

\newcommand{\Lb}{\pazocal{L}}
\usepackage{flushend}
\begin{document}

\title{Towards Explainable AI for Channel Estimation in Wireless Communications}

\author{Abdul~Karim~Gizzini,~Yahia~Medjahdi,~Ali~J.~Ghandour,~Laurent~Clavier

\thanks{
%
%
Abdul Karim Gizzini, Yahia Medjahdi, and Laurent Clavier are with the Center for Digital Systems, IMT Nord Europe, Institut Mines-Télécom, University of Lille, France (e-mail: \{abdul-karim.gizzini, yahia.medjahdi,laurent.clavier\}@imt-nord-europe.fr).

Ali J. Ghandour is with the National Center for Remote Sensing - CNRS, Lebanon (e-mail: aghandour@cnrs.edu.lb).

}
}
\markboth{} 
{}
\maketitle
\begin{abstract}

Research into 6G networks has been initiated to support a variety of critical artificial intelligence (AI) assisted applications such as autonomous driving. In such applications, AI-based decisions should be performed in a real-time manner. These decisions include resource allocation, localization, channel estimation, etc. Considering the black-box nature of existing AI-based models, it is highly challenging to understand and trust the decision-making behavior of such models. Therefore, explaining the logic behind those models through explainable AI (XAI) techniques is essential for their employment in critical applications. This manuscript proposes a novel XAI-based channel estimation (XAI-CHEST) scheme that provides detailed reasonable interpretability of the deep learning (DL) models that are employed in doubly-selective channel estimation. The aim of the proposed XAI-CHEST scheme is to identify the relevant model inputs by inducing high noise on the irrelevant ones.
As a result, the behavior of the studied DL-based channel estimators can be further analyzed and evaluated based on the generated interpretations. Simulation results show that the proposed XAI-CHEST scheme provides valid interpretations of the DL-based channel estimators for different scenarios.

\end{abstract}


\IEEEpeerreviewmaketitle
\renewcommand{\figurename}{Fig.}
\section{Introduction} \label{introduction}
It is envisioned that 6G technology will provide a new era of mass digital connectivity and automation of several advanced smart services such as autonomous driving and remote surgery~\cite{ref_xai_1}. Such services are mission critical, where high data rates, low latency, and robust communications must be established and guaranteed~\cite{ref_xai_2}. {\ac{AI}} will be a key element in future smart applications, due to its great success in providing good performance compared to conventional methods. 
However, the recent {{\ac{DL}}}-based schemes proposed for wireless communications suffer mainly from the lack of transparency and trust. In this context, the development of trustworthy {\ac{DL}}-based schemes is a crucial need in order to provide explainability for the {\ac{DL}} model decisions in a logical and understandable manner~\cite{ref_xai_1, ref_xai_3, 10292755}. Motivated by the fact that  reliable communications can be guaranteed by  accurate channel estimation,  
%
recently, {\ac{DL}} algorithms have been integrated into channel estimation~\cite{ref_DL_Chest1} to handle the limitations of conventional channel estimators represented by prior data knowledge, high complexity and statistical assumptions. These limitations lead to significant performance degradation, especially in doubly-selective environments, where the channel varies in both time and frequency. In contrast, {\ac{DL}}-based channel estimators have succeeded in improving the overall system performance, especially when used on top of low-complexity conventional estimators. This success is mainly due to the robustness, low complexity, and good generalization ability of {\ac{DL}} techniques. This generalization ability can be guaranteed by employing several techniques like the  Bayesian neural network (BNN). Unlike regular DL networks where the weights are considered as point estimates, thus, they may suffer from the overfitting problem, i.e. lack of a good generalization capability for new samples in the testing phase. BNN networks are able to tackle this issue by placing the point estimates with probability distributions, where they
measure the model uncertainty resulting from the input data and model training~\cite{9001195, 9536454}. After that, BNN applies the desired regularization according to measured uncertainty, where better generalization can be achieved.

Even though {\ac{DL}}-based channel estimators provide a good generalization and performance-complexity trade-off, they lack decision-making interpretability. Hence, they are classified as black-box {\ac{DL}} models with a main trustworthiness problem. This lack of interpretability can be addressed by developing {\ac{XAI}} schemes that increase the black-box models transparency and explain why certain decisions were made. {\ac{XAI}} schemes are able to provide reasoning and justifications of the {\ac{DL}} behavior, thus, transforming the black-box model into a white-box model and ensuring the adaptation, trustfulness and transparency of a given model. A variety of {\ac{DL}} techniques are employed in the doubly-selective channel estimation~\cite{ref_survey,ref_wi_cnn}, however, this paper considers the recently proposed {\ac{FNN}}-based channel estimators~\cite{ref_STA_DNN, ref_TRFI_DNN}.

\begin{figure}[t]
\centering
\includegraphics[width=\columnwidth] {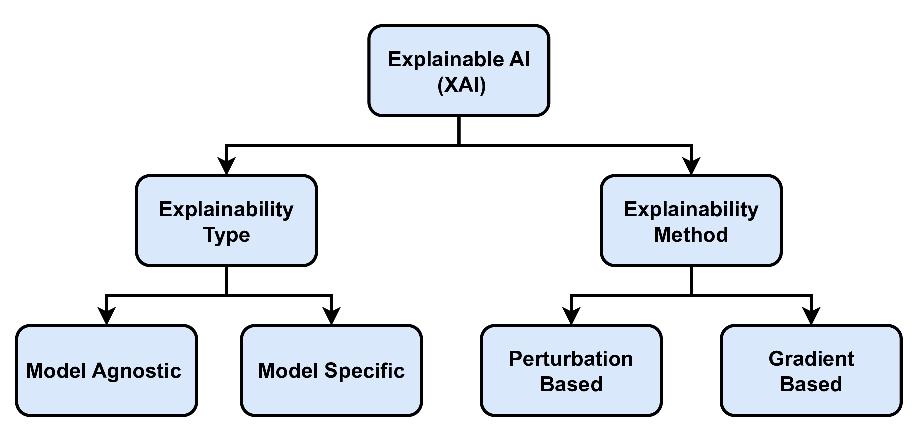}
\caption{General XAI taxonomy.}\label{fig:xai_tax}
\end{figure}

According to the {\ac{XAI}} taxonomy, as shown in Fig.~{\ref{fig:xai_tax}}, there are two main categories of XAI methods~\cite{nielsen2022robust}: (\textit{i}) perturbation-based or gradient-free methods, where the concept is to perturb input features by masking or altering their values and record the effect of these changes on the model performance and (\textit{ii}) gradient-based methods where the gradients of the output are calculated with respect to the input via back-propagation and used to estimate importance scores of the input features. 

In this context, {\ac{XAI}} schemes can be further classified into (\textit{i}) model-Agnostic: where the {\ac{XAI}} scheme does not consider the internal architecture of the model including the weights and the layers. Hence, Agnostic-based {\ac{XAI}} schemes are characterized by high flexibility since they can be applied to interpret any black-box model regardless of its architecture. (\textit{ii}) model-specific: Here the {\ac{XAI}} scheme is defined according to the architecture and parameters of a specific trained model. Thus it can not be employed for a variety of {\ac{DL}} models. 

It should be noted that \ac{XAI} schemes are used primarily in computer vision applications~\cite{koker2021u}. Procedures and methodology for adopting and deploying such schemes in the wireless communications discipline are still unclear~\cite{ref_xai_1}. In this context, this paper proposes a novel model-agnostic perturbation-based {\ac{XAI}} channel estimation (XAI-CHEST) scheme, where the interpretability of recently proposed {\ac{FNN}}-based channel estimators is investigated. The key idea is to employ an {\ac{FNN}} model, denoted as the interpretability model, to provide interpretations for the considered black-box model denoted as the utility model. The interpretability model role is to induce noise on the {\ac{FNN}} inputs without degrading the performance of the utility model. To achieve this, the interpretability model will only induce high noise on the {\ac{FNN}} inputs that are not substantial to the utility model proper functioning. As a result, the interpretability model generates a noise mask that includes the corresponding noise weight for each {\ac{FNN}} input, where the {\ac{FNN}} inputs can be classified as either relevant or irrelevant inputs. Simulation results reveal that the proposed XAI-CHEST scheme is able to accurately illustrate the behavior of the utility model, thus transforming it into a white-box model. Furthermore, we show that employing only selected relevant inputs instead of the full inputs improves the {\ac{FNN}}-based channel estimators performance. 

To the best of our knowledge, this is the first work to propose an {\ac{XAI}} scheme for physical layer applications in wireless communications, specifically, channel estimation. In summary, the contributions of this paper are threefolds

\begin{itemize}
    
    \item Proposing an XAI-CHEST scheme that provides detailed interpretability to the recently proposed {\ac{FNN}}-based channel estimators.

    \item Showing that the interpretability of the studied {\ac{FNN}} models can be achieved by inducing noise on the model inputs while maximizing their accuracy. Therefore, the inputs of the model are classified into relevant and irrelevant sets based on the induced noise mask.


    \item Providing an extensive performance evaluation of the proposed XAI-CHEST scheme in different scenarios, where we show that using only relevant inputs instead of the full set improves the performance of the studied {\ac{FNN}}-based channel estimators, as well as reducing the computational complexity. 
    
\end{itemize}

The remainder of this paper is organized as follows: Section~\ref{system_model} presents the system model. The  {\ac{DL}}-based channel estimators to be interpreted are presented in Section~\ref{soa_schemes}. Section~\ref{proposed_xai_scheme} illustrates the proposed XAI-CHEST scheme. In Section~\ref{simulation_results}, the performance of the proposed XAI-CHEST scheme in terms of \ac{BER} is analyzed. Finally, Section~\ref{conclusions} concludes the manuscript.

\section{System Model} \label{system_model}

Consider a frame consisting of $I$ {\ac{OFDM}} symbols.  The $i$-th transmitted frequency-domain {\ac{OFDM}} subcarrier $\tilde{\ma{x}}_i [k]$, is denoted by
\begin{equation}
   \tilde{\ma{x}}_i[k] = \left\{
            \begin{array}{ll}        
                \tilde{\ma{x}}_{{i,d}}[k],&\quad k \in \Kd \\
                \tilde{\ma{x}}_{{i,p}}[k],&\quad k \in \Kp \\
                0,&\quad k \in \Kn \\
            \end{array}\right.
\label{eq: xK}
\end{equation}
where $0 \leq k \leq K - 1$. The total number of used subcarriers is divided into $K_{\text{on}} = K_{d} + K_{p}$ subcarriers in addition to $K_{n}$ null guard band subcarriers, where 
$ \tilde{\ma{x}}_{{i,d}} \in \compl^{K_{d} \times 1}$ and $ \tilde{\ma{x}}_{{i,p}} \in \compl^{K_{p} \times 1}$ represent the modulated data symbols and the predefined pilot symbols allocated at a set of subcarriers denoted $\Kd$ and $\Kp$, respectively. The received frequency-domain {\ac{OFDM}} subcarrier denoted as $\tilde{\ma{y}}_{{i}}[k]$ is expressed as follows:

\begin{equation}
	\begin{split}
		\tilde{\ma{y}}_{{i}}[k] 
		&= \tilde{\ma{h}}_i[k] \tilde{\ma{x}}_i[k] + \tilde{\ma{e}}_{i}[k] + \tilde{\ma{v}}_i[k],~ k \in \Kon
	\end{split}            
	\label{eq: system_model}
\end{equation}
where $\tilde{\ma{y}}_i \in \compl^{K_{on} \times 1}$ and $\tilde{\ma{x}}_i \in \compl^{K_{on} \times 1}$ represents the received and transmitted {\ac{OFDM}} symbols, respectively. $\tilde{\ma{h}}_i \in \compl^{K_{on} \times 1}$ refers to the frequency response of the doubly-selective channel at the $i$-th {\ac{OFDM}}. ${\tilde{\ma{v}}}_i  \in \compl^{K_{on} \times 1}$ signifies the {\ac{AWGN}} of variance $\sigma^2$ and $\tilde{\ma{e}}_{i}  \in \compl^{K_{on} \times 1}$ denotes the Doppler interference derived in~\cite{ref_survey}.

\section{DL-based Channel Estimation Schemes} \label{soa_schemes}
 
Conventional channel estimators are based on preamble-based channel estimation, where the channel is estimated once at the beginning of the received frame. In this case, the estimated channel becomes outdated in high-mobility environments. Pilot subcarriers allocated within a transmitted {\ac{OFDM}} symbol are limited and cannot fully capture the doubly selective channel variations. In this context, several channel estimators are proposed in the literature to allow better channel tracking over time, where {\ac{FNN}} is applied as a post-processing on top of conventional channel estimators. In this work, the interpretability of the recently proposed {\ac{STA}}-{\ac{FNN}} and {\ac{TRFI}}-{\ac{FNN}} channel estimators is investigated.

\subsection{STA-FNN}

In the {\ac{STA}} estimator~\cite{ref_sta}, frequency- and time- domain averaging are applied on top of the {\ac{DPA}} channel estimation. We note that {\ac{DPA}} channel estimation utilizes the demapped data subcarriers of the previously received {\ac{OFDM}} symbol to estimate the channel for the current {\ac{OFDM}} symbol such that

\begin{equation}
\tilde{\ma{d}}_i[k] =  \mathfrak{D} \big( \frac{\tilde{\ma{y}}_i[k]}{\hat{\tilde{\ma{h}}}_{\text{DPA}_{i-1}}[k]}\big)
,~ \hat{\tilde{\ma{h}}}_{\text{DPA}_{0}}[k] = \hat{\tilde{\ma{h}}}_{\text{LS}}[k]
\label{eq: DPA_1}
\end{equation}
where $\mathfrak{D}(.)$ refers to the demapping operation to the nearest constellation point in accordance with the employed modulation order. $\hat{\tilde{\ma{h}}}_{\text{LS}}$ stands for the LS estimated channel at the received preambles. Thereafter, the final {\ac{DPA}} channel estimates are updated in the following manner 
\begin{equation}
\hat{\tilde{\ma{h}}}_{\text{DPA}_{i}}[k] = \frac{\tilde{\ma{y}}_i[k]}{\tilde{\ma{d}}_i[k]}
\label{eq: DPA_2}
\end{equation}

Finally, frequency- and time- domain averaging are implemented as follows:

\begin{equation}
\hat{\tilde{\ma{h}}}_{\text{FD}_{i}}[k] = \sum_{\lambda = -\beta}^{\lambda = \beta} \omega_{\lambda} \hat{\tilde{\ma{h}}}_{\text{DPA}_{i}}[k + \lambda], ~ \omega_{\lambda} = \frac{1}{2\beta+1}
\label{eq: STA_4}
\end{equation}
\begin{equation}
\hat{\tilde{\ma{h}}}_{\text{STA}_{i}}[k] = (1 - \frac{1}{\alpha})  \hat{\tilde{\ma{h}}}_{\text{STA}_{i-1}}[k] + \frac{1}{\alpha}\hat{\tilde{\ma{h}}}_{\text{FD}_{i}}[k]
\label{eq: STA_5}
\end{equation}
{\ac{STA}} estimator performs well in low {\ac{SNR}} region. However, it suffers from a considerable error floor in high {\ac{SNR}} regions due to: (\textit{i}) large {\ac{DPA}} demapping error and (\textit{ii}) fixed frequency and time averaging coefficients $\alpha = \beta = 2$. Therefore, {\ac{STA}} channel estimator suffers from a significant performance degradation in real-case scenarios due to the high doubly-selective channel variations. As a workaround, {\ac{FNN}} is utilized as a nonlinear post-processing unit following {\ac{STA}}~\cite{ref_STA_DNN}. STA-FNN is able to better capture the time-frequency correlations of the channel samples, in addition to correcting the conventional {\ac{STA}} estimation error.

\begin{figure*}[t]
\centering
\includegraphics[width=1.7\columnwidth, height=4cm]{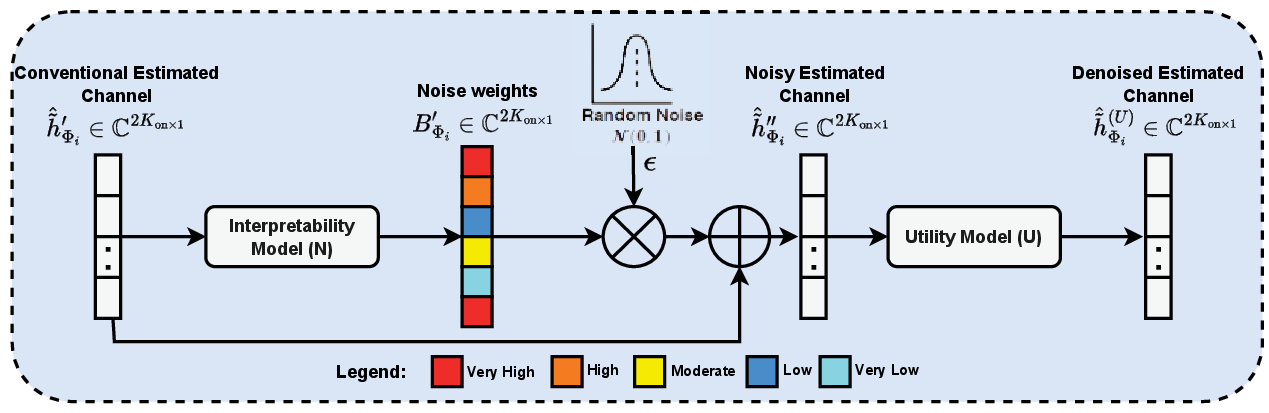}
\caption{Block diagram of the proposed XAI-CHEST scheme.}\label{fig:proposed_xai_scheme}
\end{figure*}

\subsection{TRFI-FNN}
{\ac{TRFI}} estimation scheme~\cite{ref_TRFI} is another methodology used for improving the {\ac{DPA}} estimation in~\eqref{eq: DPA_2}. Assuming that the time correlation of the channel response between two adjacent {\ac{OFDM}} symbols is high, {\ac{TRFI}} defines two sets of subcarriers such that: (\textit{i}) {$\RS_{i}$} set that includes the reliable subcarriers indices,  and (\textit{ii}) {$\URS_{i}$} set which contains the unreliable subcarriers indices. The estimated channels for {$\URS_{i}$} are then interpolated using the {$\RS_{i}$} channel estimates by means of the frequency-domain cubic interpolation\footnote{The reliable/unreliable subcarriers are classified by TRFI channel estimator, while relevant/irrelevant subcarriers are classified by the proposed XAI scheme.}.  {\ac{TRFI}} outperforms {\ac{STA}}, especially in high {\ac{SNR}} region. However, {\ac{TRFI}} still suffers from demapping and interpolation errors, as the number of \ac{RS} subcarriers is inversely proportional to the channel frequency selectivity. Therefore, applying cubic interpolation with only a few \ac{RS} subcarriers degrades the performance. Therefore, an optimized \ac{FNN} architecture is coined in~\cite{ref_TRFI_DNN} with $\hat{\tilde{\ma{h}}}_{\text{TRFI}_{i}}[k]$ as an input instead of  $\hat{\tilde{\ma{h}}}_{\text{STA}_{i}}[k]$. {\ac{TRFI}}-{\ac{FNN}} corrects the cubic interpolation error, thus leading to improved performance in high {\ac{SNR}} regions. Since real-valued {\ac{FNN}} model is considered, the conventional estimated channel should be converted from complex to real domain by stacking the real and imaginary parts of {\ac{FNN}} input vector, where $\hat{\tilde{\ma{h}}}^{\prime}_{\Phi_{i}} \in \mathbb{R}^{2 K_{\text{on}} \times 1}$ denotes the converted estimated channel vector, such that 

\begin{equation}
\hat{\tilde{\ma{h}}}^{\prime}_{\Phi_{i}} = [\Re(\hat{\tilde{\ma{h}}}_{\Phi_{i}}) ; \Im(\hat{\tilde{\ma{h}}}_{\Phi_{i}})]
\label{eq:complextoreal}
\end{equation}
where $\hat{\tilde{\ma{h}}}_{\Phi_{i}} \in \compl^{K_{\text{on}} \times 1}$ denotes the initially estimated channel and $\Phi \in [\text{STA,~TRFI}]$ refers to the estimation scheme.

\section{Proposed XAI-CHEST Scheme} \label{proposed_xai_scheme}

As discussed in the introduction, the interpretability of the black box model could be achieved either by studying its internal architecture or by externally analyzing the input-output relation. In this context, the proposed XAI-CHEST scheme provides a perturbation-based external interpretability of the black box channel estimation model, where the model inputs are classified as relevant and irrelevant. The main intuition is that if a subcarrier is relevant for the decision-making of a trained black box model, then adding noise with high weight to this subcarrier would negatively impact the accuracy of the model and vice-versa. Thus, it is expected that considering only the relevant subcarriers as model inputs would improve channel estimation performance. The methodology of the proposed XAI scheme is explained as follows.

Let $U$ be the utility model with parameters $\theta_U$. The objective is to provide interpretability for the decisions taken by the $U$ model that refers to the black box model, i.e., STA-FNN or TRFI-FNN, in our case. We also define the interpretability noise model $N$, with parameters $\theta_N$, whose purpose is to compute the value of the induced noise weight to each subcarrier within the {\ac{FNN}} input vector, i.e. $\hat{\tilde{\ma{h}}}_{\text{STA}_{i}}$ or $\hat{\tilde{\ma{h}}}_{\text{TRFI}_{i}}$. The key idea lies in the customized loss function of the $N$ model that will adjust the induced noise while simultaneously maximizing the performance of the $U$ model. We note that both $U$ and $N$ models have the same {\ac{FNN}} architecture, and the $U$ model is trained prior to the XAI processing, i.e., the weights of the $U$ model are frozen.

Let $\hat{\tilde{\ma{h}}}^{\prime}_{\Phi_{i}} \in \mathbb{R}^{2 K_{\text{on}} \times 1}$ be the interpretability model input, where the latter produces a mask $B_{\Phi_{i}}^{\prime} \in \mathbb{R}^{2 K_{\text{on}} \times 1}$ that can be represented as follows

\begin{equation}
   B_{\Phi_{i}}^{\prime} = N(\hat{\tilde{\ma{h}}}^{\prime}_{\Phi_{i}}, \theta_{N})
\end{equation}
where $ B_{\Phi_{i}}^{\prime} \in [0,1]$ determines the weight of noise applied to each element in $\hat{\tilde{\ma{h}}}^{\prime}_{\Phi_{i}}$. We note that the scaling of $ B_{\Phi_{i}}^{\prime}$ is achieved by using the sigmoid activation function. Based on $B_{\Phi_{i}}^{\prime}$, the conventional estimated channel vector including the induced noise  $\hat{\tilde{\ma{h}}}^{\prime\prime}_{\Phi_{i}}$ can be expressed such that

\begin{equation}
  \hat{\tilde{\ma{h}}}^{\prime\prime}_{\Phi_{i}} = \hat{\tilde{\ma{h}}}^{\prime}_{\Phi_{i}} + B_{\Phi_{i}}^{\prime} \epsilon
\end{equation}

Here, $\epsilon \sim  \mathcal{N} (0,1)$ denotes the random noise sampled from the standard normal distribution. After that, $\hat{\tilde{\ma{h}}}^{\prime\prime}_{\Phi_{i}}$ is fed as input to the frozen utility model $U$, i.e, {\ac{STA}}-{\ac{FNN}} or {\ac{TRFI}}-{\ac{FNN}}, such that

\begin{equation}
   \hat{\tilde{\ma{h}}}^{(\text{U})}_{\Phi_{i}} = U(\hat{\tilde{\ma{h}}}^{\prime\prime}_{\Phi_{i}}, \theta_{U})
\end{equation}

The aim of training the interpretability model is to minimize the following customized loss function:

\begin{equation}
\Lb_{N} = \min_{\theta_{N}}  \big[ \Lb_{U} - \lambda \log(B_{\Phi_{i}}^{\prime}) \big]
\label{eq:lossn}    
\end{equation}


The first term in Eq. ~\eqref{eq:lossn} is the loss function of the utility model $U$ that represents\ac{MSE} between the initially estimated channel $\hat{\tilde{\ma{h}}}^{\prime}_{\Phi_{i}}$ and the true channel ${\tilde{\ma{h}}}_{\Phi_{i}}$, such that:

\begin{equation}
\Lb_{U} = \frac{1}{N_{tr}}. \sum_{i = 1}^{N_{tr}} \big( {\tilde{\ma{h}}}_{\Phi_{i}} - \hat{\tilde{\ma{h}}}^{\prime}_{\Phi_{i}} \big)^{2}
\label{eq:lossu}
\end{equation}
where $N_{tr}$ is the number of training samples used. The second term in~\eqref{eq:lossn} boosts the growth of the noise weight for each element in $\hat{\tilde{\ma{h}}}^{\prime\prime}_{\Phi_{i}}$ so that relevant and irrelevant subcarriers can be identified. We would like to mention that the training of the interpretability model $N$ is performed once according to the channel model used. In the testing phase, the obtained $B_{\Phi_{i}}^{\prime}$ includes the noise weights for the real and imaginary parts of each input subcarrier. $B_{\Phi_{i}}^{\prime}$ is scaled back to $B_{\Phi_{i}} \in \mathbb{R}^{K_{\text{on}} \times 1}$, where the noise weight of the real and imaginary parts for each subcarrier are averaged. The motivation behind this averaging lies in the fact that it is noticed during the training phase that the interpretability model produces almost the same noise weight for the real and imaging parts of each subcarrier within $\hat{\tilde{\ma{h}}}^{\prime}_{\Phi_{i}}$. \figref{fig:proposed_xai_scheme} shows the block diagram of the proposed XAI-CHEST scheme.

Algorithm~{\ref{algo:noise_training}} illustrates the training process of the interpretability model (N). $\hat{\tilde{\ma{H}}}^{\prime}_{\Phi} \in \mathbb{R}^{2 K_{\text{on}} \times I_{\text{tr}}} $ and ${\tilde{\ma{H}}}_{\Phi} \in \mathbb{R}^{2 K_{\text{on}} \times I_{\text{tr}}} $ denote the training dataset pairs of the conventional estimated channels and the true ones, where $I_{\text{tr}}$ is the size of the training dataset.

\begin{algorithm}
\caption{Interpretability model (N) training}
\begin{algorithmic} 
\REQUIRE 
 Estimated channel: $\hat{\tilde{\ma{H}}}^{\prime}_{\Phi}$, true channel: ${\tilde{\ma{H}}}_{\Phi}$, learning rate: $\eta$, trained utility model $U$ with
parameters $\theta_U$
\ENSURE Trained interpretability model (N) with parameters $\theta_N$
\WHILE{not converged}
\FOR{$\hat{\tilde{\ma{h}}}^{\prime}_{\Phi_{i}} \in \hat{\tilde{\ma{H}}}^{\prime}_{\Phi_{i}}$, ${\tilde{\ma{h}}}_{\Phi_{i}} \in {\tilde{\ma{H}}}_{\Phi_{i}}$}    
\STATE { $B_{\Phi_{i}}^{\prime}$ $\gets$ $N(\hat{\tilde{\ma{h}}}^{\prime}_{\Phi_{i}}, \theta_{N})$}
\STATE {$\epsilon$ $\gets$ $\mathcal{N} (0,1)$}
\STATE {$\hat{\tilde{\ma{h}}}^{\prime\prime}_{\Phi_{i}}$ $\gets$ $\hat{\tilde{\ma{h}}}^{\prime}_{\Phi_{i}} + B_{\Phi_{i}}^{\prime} \epsilon$}
\STATE {$\hat{\tilde{\ma{h}}}^{(\text{U})}_{\Phi_{i}}$ $\gets$ $ U(\hat{\tilde{\ma{h}}}^{\prime\prime}_{\Phi_{i}}, \theta_{U})$}
\STATE{ $\Lb_{U}$ $\gets$ $ \text{MSE} \big( {\tilde{\ma{h}}}_{\Phi_{i}} - \hat{\tilde{\ma{h}}}^{(\text{U})}_{\Phi_{i}} 
\big)$}
\STATE{ $\Lb_{N}$ $\gets$ $\Lb_{U}$ $ - \lambda \log(B_{\Phi_{i}}^{\prime})$}
\STATE {$\theta_N$ $\gets$ $\theta_N +  \eta \frac{\partial \Lb_{N} }{\partial \theta_N }$ } 
\ENDFOR
\ENDWHILE
\end{algorithmic}
\label{algo:noise_training}
\end{algorithm}

\section{Simulation Results} \label{simulation_results}

\begin{figure*}[t]
	\setlength{\abovecaptionskip}{6pt plus 3pt minus 2pt}
	\centering
	\includegraphics[width=2\columnwidth]{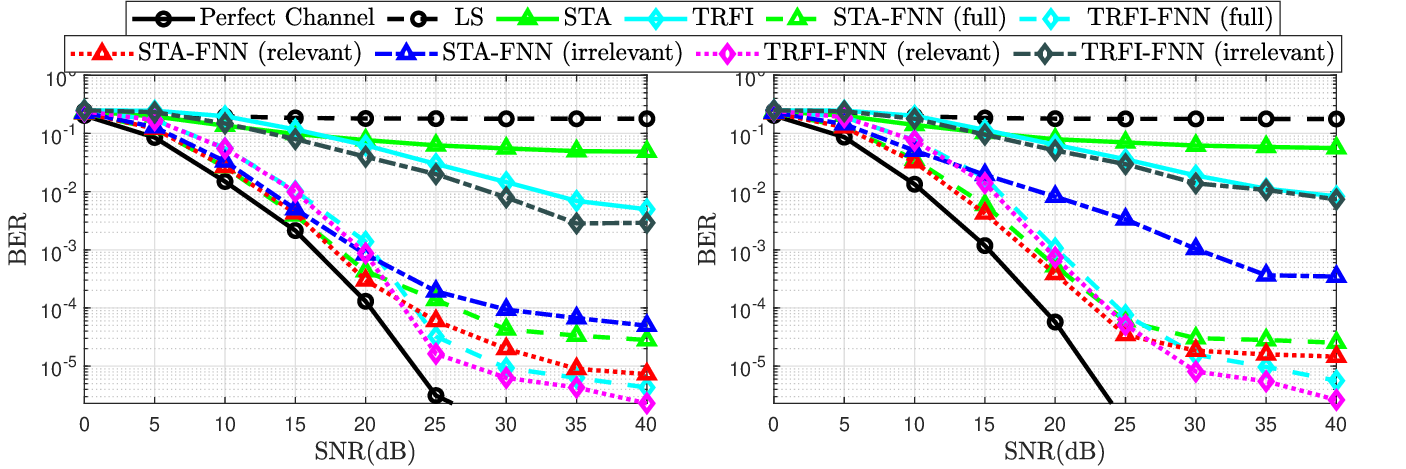}
	\subfloat[\label{fig:LF_BER} Low frequency-selective vehicular channel model.]{\hspace{.5\linewidth}}
	\subfloat[\label{fig:HF_BER} High frequency-selective vehicular channel model.]{\hspace{.5\linewidth}} 
\caption{BER Performance for high mobility vehicular channel models.}
	\label{fig:BER}
\end{figure*}

\begin{figure*}[t]
	\setlength{\abovecaptionskip}{6pt plus 3pt minus 2pt}
	\centering
	\includegraphics[width=2\columnwidth]{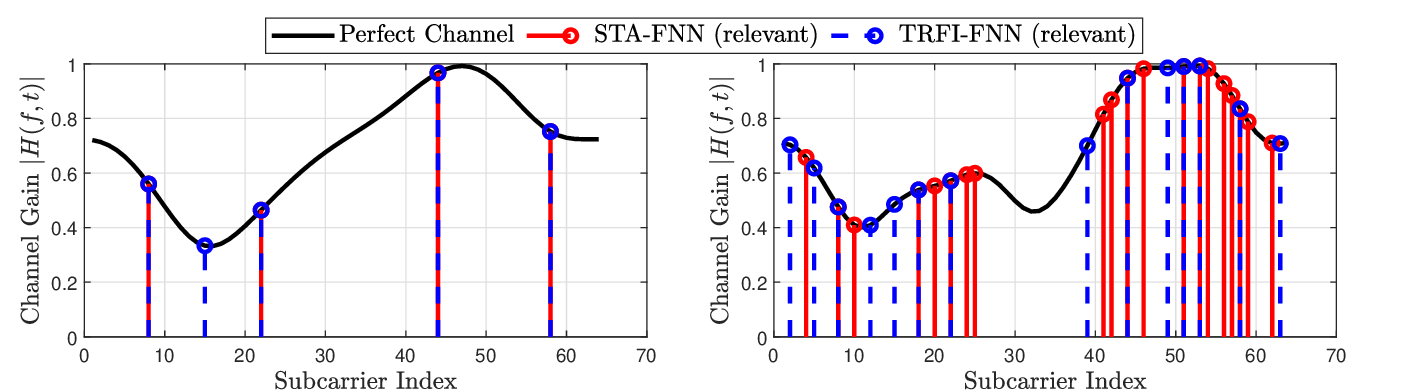}
	\subfloat[\label{fig:LF_channel_interpolation} Low frequency-selective vehicular channel model.]{\hspace{.5\linewidth}}
	\subfloat[\label{fig:HF_channel_interpolation} High frequency-selective vehicular channel model.]{\hspace{.5\linewidth}} 
\caption{Selected relevant subcarriers by the proposed interpretability model.}
\label{fig:channel_interpolation}
\end{figure*}

This section illustrates the performance evaluation of the proposed XAI-CHEST scheme, where {\ac{BER}} performance of STA-FNN and TRFI-FNN estimators is analyzed taking into consideration full, relevant, and irrelevant subcarriers. We note that the noise thresholds used to classify the subcarriers are empirically chosen and defined as $N_{\text{TRFI-th}} = 0.6$ and  $N_{\text{STA-th}} = 0.3$, where the systematic fine-tuning of them is kept as future work. This means that all the subcarriers given a noise weight greater than the defined threshold are classified as irrelevant subcarriers and vice-versa.  


BER is evaluated using two vehicular channel models~\cite{r19} as shown in Table~{\ref{tb:VCMC}}: (\textit{i}) Low-frequency selectivity, where \ac{VTV-US}  is employed. (\textit{ii}) High-frequency selectivity, where \ac{VTI-US} is considered. In both scenarios, Doppler frequency $f_d = 1000$ Hz is considered. 
Both the interpretability and utility models are trained using a $100, 000$ OFDM symbols dataset, splitted into $80\%$ training and $20\%$ testing. ADAM optimizer is used with a learning rate $lr = 0.001$ with batch size equals $128$ for $500$ epoch. Simulation parameters are based on the IEEE 802.11p standard~{\cite{ref_survey}}, where the comb pilot allocation is used so that $K_{p} = 4$, $K_{d} = 48$, $K_{n} = 12$, and $I = 50$. QPSK modulation is considered with SNR ranges $[0,5,\dots,40]$ dB.

\begin{table*}[t]
\renewcommand{\arraystretch}{1.6}
\centering
\caption{Characteristics of the employed channel models following Jake's Doppler spectrum.}
\label{tb:VCMC}
\begin{tabular}{|c|c|c|}
\hline
\textbf{Channel model} & \textbf{Average path gains {[}dB{]}}                                                                                       & \textbf{Path delays {[}ns{]}}                                                                            \\ \hline
VTV-US                 & \begin{tabular}[c]{@{}c@{}}{[}0, 0, -10, -10, -10, -17.8, -17.8, -17.8, -21.1, -21.1, -26.3, -26.3{]}\end{tabular}     & \begin{tabular}[c]{@{}c@{}}{[}0, 1, 100, 101, 102, 200, 201,202, 300, 301,400, 401{]}\end{tabular} \\ \hline
VTI-US             & \begin{tabular}[c]{@{}c@{}}{[}0, 0, -9.3, -9.3, -14, -14, -18, -18, -19.4,-24.9, -27.5, -29.8{]}\end{tabular} & \begin{tabular}[c]{@{}c@{}}{[}0, 1, 100, 101, 200, 201, 300,301, 400, 500, 600, 700{]}\end{tabular} \\ \hline
\end{tabular}
\end{table*}

\figref{fig:BER} depicts the {\ac{BER}} performance of the STA-FNN and TRFI-FNN channel estimation schemes studied. It is clearly shown that employing only the LS channel estimation suffers from severe performance degradation, as the channel becomes outdated among the received OFDM symbols. Moreover, limited improvement can be achieved with the conventional STA and TRFI channel estimation scheme due to the enlarged DPA demapping error. This is attributed to the fact that conventional STA estimation outperforms TRFI in low SNR regions because of frequency and time averaging operations that can alleviate the impact of noise and demapping error in low SNR regions. However, the averaging operations are not useful in high SNR regions since the impact of noise is low, and the STA averaging coefficients are fixed. Hence, the TRFI channel estimation is more accurate in high SNR regions. As observed, FNN can implicitly learn the channel correlations apart from preventing a high demapping error arising from conventional estimation where STA-FNN and TRFI-FNN significantly outperform conventional STA and TRFI estimators.

In order to evaluate the performance of the proposed XAI-CHEST scheme, both STA-FNN and TRFI-FNN estimators are implemented in three different configurations: (\textit{i}) Full where $\hat{\tilde{\ma{h}}}^{\prime}_{\Phi_{i}}  \in \mathbb{R}^{2 K_{\text{on}} \times 1}$ includes all sub-carriers as an FNN input. (\textit{ii}) Relevant where only relevant subcarriers experiencing noise below the defined $N_{\Phi\text{-th}}$ are fed to the FNN. (\textit{ii}) Irrelevant where irrelevant subcarriers recording noise weights above the defined threshold are considered. It is worth mentioning that when the relevant subcarriers are used instead of the full subcarriers, STA-FNN and TRFI-FNN perform better and $2$ dB and $1$ dB gain are achieved in terms of SNR for BER $= 10^{-4}$, respectively. On the contrary, significant performance degradation can be noticed for both STA-FNN and TRFI-NN estimators when irrelevant subcarriers are used, even though the number of irrelevant subcarriers is greater than the number of relevant subcarriers\footnote{BER is computed over all the data subcarriers within the received OFDM symbols. The difference lies in the size of the FNN input vector that includes the relevant or irrelevant subcarriers.}. 

In order to further validate the recorded BER results, the position of relevant subcarriers selected by the interpretability model is analyzed, as illustrated in~\figref{fig:channel_interpolation}. It is clearly shown that the relevant subcarriers experiencing low noise induced by the interpretability model are distributed among the sharp channel variations. This reveals that it is not necessary for the FNN model to have the full $K_{on} = 52$ subcarriers as an input. It can perform better by choosing the appropriate subcarriers to capture the main channel variation among the subcarriers. Moreover, the number of selected relevant subcarriers increases with the increase in frequency selectivity. In high-frequency selectivity, the channel variation across the subcarriers becomes more challenging, and thus more subcarriers are required by the FNN to accurately estimate the channel. Moreover, it can be noticed that as the accuracy of the initial estimation increases, the number of selected relevant subcarriers decreases, where STA-FNN requires more relevant subcarriers than TRFI-FNN. This means that the interpretability model is able to induce more noise weight to the irrelevant subcarriers within TRFI-FNN since it is more accurate, whereas less noise is induced to the STA-FNN input since it is already noisy. Finally, fine-tuning the noise threshold can be achieved by studying the characteristics of the channel model employed and the BER requirement. However, in all cases, using the proposed XAI-CHEST scheme provides a reasonable interpretation of the behavior of the employed FNN black-box model. In addition, selecting the relevant FNN inputs leads to improving the BER performance as well as decreasing the computational complexity due to optimizing the 
FNN input and the FNN architecture accordingly.

\acresetall
\section{Conclusion} \label{conclusions}

Building trust and transparency of the AI-based solution is a critical issue in future 6G communications. The need to provide reasonable interpretability of the black box AI models is a must. In this context, an XAI-CHEST scheme is proposed where the interpretability of the recently proposed FNN-based channel estimation schemes is thoroughly investigated. The proposed XAI-CHEST scheme aims to induce higher noise weights on the irrelevant FNN input subcarriers, thus, enabling a trustworthy, optimized, and robust channel estimation. Simulation results reveal that employing only the relevant subcarriers within the channel estimation leads to a significant improvement in the BER performance while optimizing the FNN input. Therefore, ensuring that FNN models are able to learn the most relevant features that maximize the channel estimation accuracy, as well as, reducing the computational complexity of the FNN-based channel estimators. 
 It is worth mentioning that the proposed XAI methodology can be easily adapted to any application other than channel estimation. As a future perspective, performance improvement using several pilot pattern designs according to the generated noise weights will be studied. In addition, the possibility of optimizing the FNN architecture based on the reduced input size as well as the required inference time will be investigated.

\ifCLASSOPTIONcaptionsoff
  \newpage
\fi
\bibliographystyle{IEEEtran}
\bibliography{ref}

\end{document}